\pdfoutput=1

\documentclass[11pt]{article}

\usepackage[final]{acl}
\usepackage{booktabs}
\usepackage{subcaption} 
\usepackage{multirow}
\usepackage[table]{xcolor}
\usepackage{times}
\usepackage{latexsym}
\usepackage{microtype}
\usepackage{inconsolata}

\usepackage{array}
\usepackage{float}

\usepackage[T1]{fontenc}

\usepackage[utf8]{inputenc}

\usepackage{amsmath}

\usepackage{graphicx}
\usepackage{enumitem}

\title{Anatomy of a Feeling: Narrating Embodied Emotions via Large Vision-Language Models}

\author{Mohammad Saim, Phan Anh Duong, Cat Luong, Aniket Bhanderi, Tianyu Jiang \\
        University of Cincinnati \\
        \texttt{saimmd@mail.uc.edu, tianyu.jiang@uc.edu}}

\begin{document}
\maketitle
\begin{abstract}
The embodiment of emotional reactions from body parts contains rich information about our affective experiences. We propose a framework that utilizes state-of-the-art large vision-language models (LVLMs) to generate \textbf{E}mbodied \textbf{L}VLM \textbf{E}motion \textbf{Na}rratives \textbf{(ELENA)}. These are well-defined, multi-layered text outputs, primarily comprising descriptions that focus on the salient body parts involved in emotional reactions. We also employ attention maps and observe that contemporary models exhibit a persistent bias towards the facial region. Despite this limitation, we observe that our employed framework can effectively recognize embodied emotions in face-masked images, outperforming baselines without any fine-tuning. ELENA opens a new trajectory for embodied emotion analysis across the modality of vision and enriches modeling in an affect-aware setting.
\end{abstract}

\section{Introduction}

Emotion recognition has been widely studied in both natural language processing and computer vision, with applications ranging from sentiment analysis to human-robot interaction. While facial expressions are often considered the primary channel for conveying emotion, they are not always reliable---for example, when faces are distant, obscured, or deliberately masked. In such cases, the body (torso and limbs) frequently communicates emotion more clearly than the face. For instance, observers often struggle to distinguish between positive and negative feelings from isolated facial expressions, but can do so accurately from bodily cues \cite{aviezer2012body}. These findings highlight that bodily reactions are central to how emotions are expressed and perceived \cite{fuchs2014embodied}; yet, this important dimension remains underexplored in much of the current emotion recognition research.

\begin{figure}[h]
  \includegraphics[width=\columnwidth]{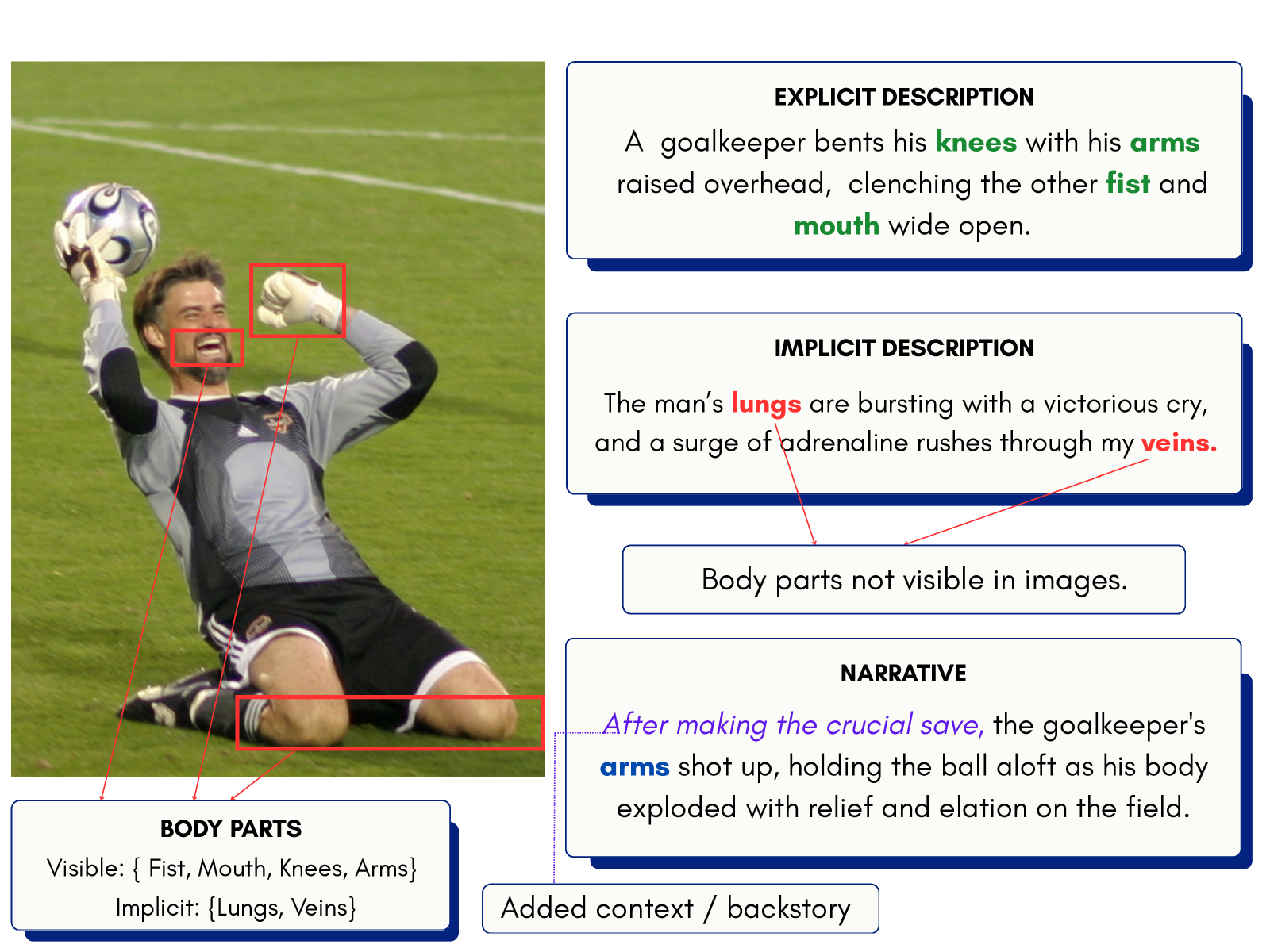}
  \caption{Illustration of embodied emotion for multi-modal analysis.}
  \label{fig:ill_example}
\end{figure}

The concept of \textit{Embodied Emotion} provides a richer framework for understanding the affective states and making holistic use of all bodily parts for emotion recognition. Rooted in psychology and cognitive science~\citep{ee_niendenthal, barsalou2008grounded}, embodiment theories argue that emotions are not purely mental phenomena but are deeply intertwined with bodily experience and physical response. In parallel, computational work has shown that embodied signals are also crucial for interpreting emotions in text, where bodily actions and sensations often convey affect beyond lexical sentiment~\citep{zhuang-etal-2024-heart, duong-etal-2025-cheer}. Together, these insights suggest that posture, gesture, and physiological change are integral to emotional meaning~\cite{barsalou2005situating, niedenthal2014embodied}. Empirical studies further show systematic correlation between specific posture and mood---for example, an upright stance can indicate elevated mood, whereas a slumped shoulder correlates with depression and low arousal \cite{wilkes2017upright}.

In this work, we propose to bridge the gap of embodied emotions research with vision modality by utilizing prevalent Large Vision-Language Models (LVLMs) and their capabilities. While prior approaches to the task of emotion recognition have often relied on models trained and fine-tuned on a variety of features \citep{tzirakis2017end2endmodel, luna2021multimodal}, our work utilizes LVLMs in a zero-shot setting to output multi-layered textual descriptions for embodied emotions, as shown in Figure~\ref{fig:ill_example}. These models do not receive any task-specific training, but are guided by carefully crafted instructions that elicit the underlying emotional manifestations in images. As far as we are concerned, this is the first work in the area of multi-modal analysis for embodied emotions. We release the code and outputs of ELENA publicly.\footnote{\url{https://github.com/cincynlp/ELENA}} We highlight our contributions as below:

\begin{enumerate}[itemsep=3pt]
    \item We introduce ELENA, a novel framework that utilizes structured prompting to guide LVLMs in generating multi-layered narratives of embodied emotion. Our approach moves beyond simple classification to a more explanatory form of affective analysis, and serves as a novel tool for qualitatively probing a model's emotional reasoning.
    \item Through systematic face-masking methodology and attention analysis, we provide the empirical evidence that LVLMs exhibit a critical \textit{attention adaptation failure} when recognizing embodied emotion---they do not redirect visual focus to informative bodily regions when facial details are masked, exposing a fundamental vulnerability in visual reasoning.
    \item We show that the framework can effectively overcome limitations of LVLMs, achieving credible performance improvements.  It suggests that while the model's autonomous visual systems are brittle, their latent knowledge of non-facial emotional signals can be effectively elicited with targeted guidance.
\end{enumerate}

\begin{figure*}[t]
\includegraphics[width=0.99\linewidth]{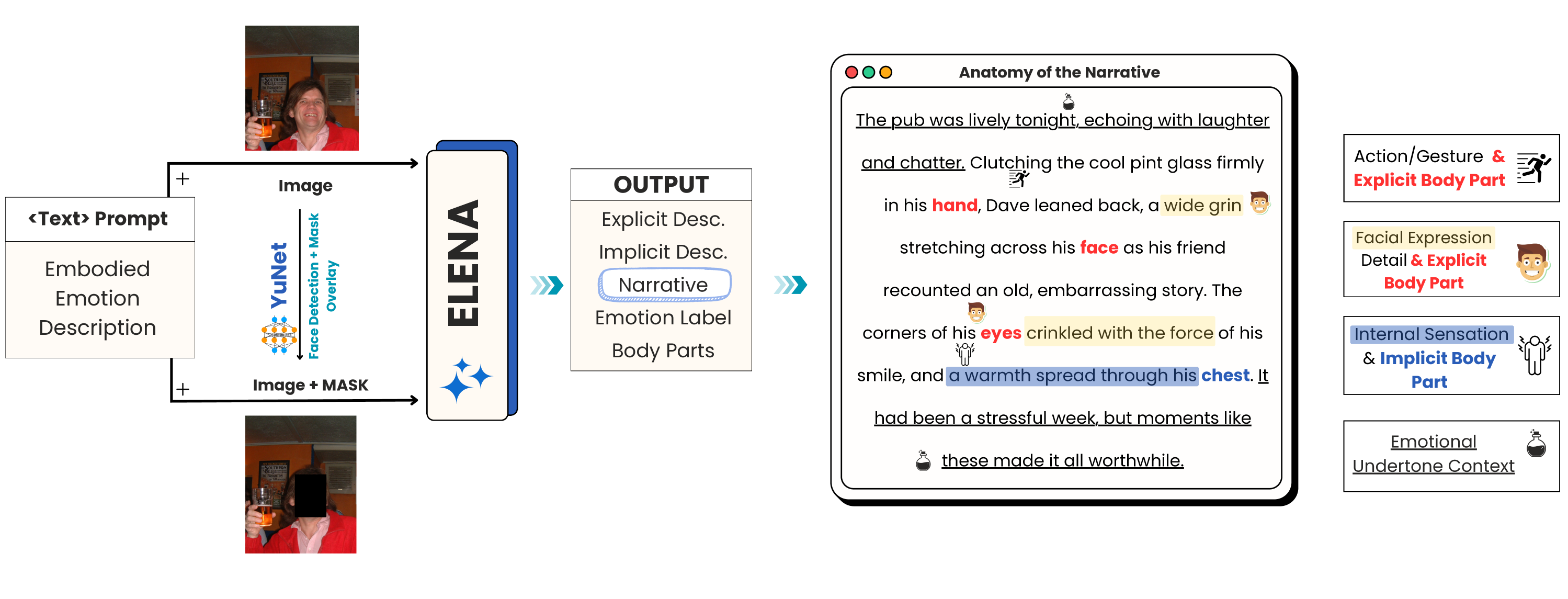}

\caption{ELENA framework architecture for embodied emotion analysis. The system utilizes LVLMs in zero-shot settings to process structured prompts with images (masked or unmasked), generating all outputs in a single forward pass: emotion labels, explicit descriptions, implicit descriptions, and body parts. Narratives are formed by combining explicit and implicit descriptions, with the right panel illustrating the anatomization of a narrative into its embodied components.}
\label{fig:pipeline}
\end{figure*}

\section{Related Work}

Emotion understanding is a long-standing task in NLP, rooted in the field of affective computing research \citep{firdaus-etal-2020-MEISD, zhuang-jiang-riloff-2020-affective, deng2021survey-TER, zheng2023facial}. Psychology and neuroscience theories have long argued that bodily states (e.g., posture and physiological reactions) are integral to emotions \citep{ ee_niendenthal, barsalou2008grounded, li-etal-2021-past-present}.

Prior NLP work has begun to consider the physical expressions of emotion in text. \citet{kim-klinger-2019-analysis} analyzed how authors depict characters’ body movements and sensations (e.g., frowning and heart pounding) to imply emotion, and \citet{casel-etal-2021-emotion} identified emotion-related components based on a cognitive model. 
The rise of large language models (LLMs) has prompted a re-examination of emotion understanding in NLP \citep{Liu_2024, sabour-etal-2024-emobench}. However, purely text-based models may overlook nonverbal signals that are critical to emotional expressions \citep{lian2025affectgpt}. 

Vision-Language Models (VLMs) extend emotion understanding with visual context and commonsense reasoning~\citep{zhang-etal-2024-visual,xenos2024vllmsprovidebettercontext}, yet evaluations on evoked emotions reveal biases towards specific categories and prompt sensitivity~\citep{bhattacharyya-wang-2025-evaluating}. Visual affective computing remains dominated by facial cues~\citep{wang2024surveyfacialexpressionrecognition}. In parallel, body-centric approaches model affect via posture and gesture features~\citep{noroozi2018survey} and skeleton-based relations among body parts \citep{8803460,wu2024upper,lu2025understanding}, but progress is limited by scarce, richly annotated datasets. Departing from these facial and skeleton-centric pipelines, we address embodied emotion in images by instructing LVLMs to generate structured, body-part-grounded narratives alongside an Ekman emotion label, and we explicitly disentangle facial versus non-facial evidence via face masking and attention-map diagnostics.

The recently introduced CHEER dataset \citep{zhuang-etal-2024-heart} represents a first step in this direction---it contains 7,300 instances of body part mentions in narratives, labeled for whether they signal an underlying emotion, e.g., “her hands trembled with fear.” This work was carried forward by \citet{duong-etal-2025-cheer}, who converted the previous binary task to Ekman's six emotion categories, along with improved metrics compared to supervised methods using best-worst scaling. However, the task of embodied recognition remains understudied in vision-language models, as well as the evaluation of the influence of facial parts of the body on emotional expression. Inferring affect purely from bodily posture, movements, and physiological state in images remains an open challenge, which motivates the focus of our study.
\section{Methodology}
We propose a zero-shot structured prompting approach that utilizes large vision-language models to systematically interpret emotions through bodily expressions. Our methodology includes detailed prompt design, face-masking experiments, attention analysis, and comparative evaluations to robustly assess the models' capability in recognizing embodied emotions.

\subsection{Datasets and Usage Scheme}

We perform our task on three image datasets:

\textbf{BESST} (Bochum Emotional Stimulus Set)~\citep{thoma2013besst} consists of 1,129 (565 frontal-body + 564 averted-body) high-quality images collected in a lab-controlled environment. An example is shown in  Appendix \ref{sec:adddatasets} (Figure \ref{fig:besst}). Participants for this research were instructed to enact a specific emotion by posing as if they were experiencing it, producing explicit bodily expressions. Each person has their face masked and photographed in both frontal and averted views. BESST includes annotations aligned with Ekman’s six basic emotions \cite{ekman1992there}, along with an additional \textit{Neutral} category.

\textbf{HECO} (Human Emotions in Context)~\citep{yang2022emotion} dataset contains 9,385 images with 19,781 annotated agents depicting people in natural settings. Emotions are labeled using the six Ekman categories, plus two additional positive states: \textit{Peace} and \textit{Excitement}. For consistency, we remap \textit{Peace} to \textit{Neutral} and \textit{Excitement} to \textit{Happiness}, aligning HECO with the seven-category taxonomy used in BESST.

\textbf{EMOTIC}~\citep{kosti2019context} dataset includes 23,571 images with 34,320 annotated people in everyday scenarios, annotated with multi-label emotion tags drawn from a fine-grained set of 26 categories. To enable consistent, cross-dataset comparisons, we map EMOTIC’s labels to the same seven-category Ekman-based scheme used for BESST and HECO. This normalization ensures a more equitable, ``apples-to-apples'' evaluation of LVLM performance across datasets. The mapping details and justification are provided in Table~\ref{tab:emotion_mapping}, Appendix~\ref{sec:addres}. For our experiments, we further simplify the evaluation by selecting the dominant emotion per instance, determined by matching gold body bounding boxes and masked face coordinates with high confidence overlap.

\subsection{Framework Overview}

The ELENA framework is illustrated in Figure \ref{fig:pipeline}. The pipeline begins with an input image (either masked or unmasked) that an LVLM processes with a text input grounded in the definition of embodied emotion. The input comprises a request for a set of varied features related to the person’s emotional state and body parts, for the output (Table \ref{tab:output_format}).  In this study, we focus on a salient person---the most notable individual in each image. The model predicts a single label, thereby maintaining uniformity across all images. However, the framework can be modified for multi-label output. The complete response structure is presented in Table \ref{tab:output_format}. Our framework can be formalized as:
\begin{equation}
\begin{aligned}
\text{ELENA}: I &\rightarrow (L, E_e, E_i, N, B)
\end{aligned}
\end{equation}
where \textit{I} represents the input image, $L \in$ \{\textit{Happiness, Sadness, Anger, Fear, Disgust, Surprise, Neutral}\} is the predicted emotion label following Ekman's taxonomy, $E_e$ denotes embodied descriptions focusing on visible body parts, $E_i$ represents implicit descriptions capturing internal sensations, and $B$ contains the set of extracted body parts. \textit{N} is the narrative that we instruct the model to produce, weaving these embodied cues with scene context.

\begin{table}[ht] 
\resizebox{0.99\linewidth}{!}{
\begin{tabular}{p{0.3\linewidth} p{0.6\linewidth}}
\toprule
\textbf{Output} & \textbf{Description} \\
\midrule
Label \textit{L} & Single emotion (Ekman's six labels + Neutral). \\
\addlinespace
Explicit \textit{$E_e$ }  & Visible body parts and their emotional expression. \\
\addlinespace
Implicit \textit{$E_i$ } & Internal sensations and body parts which are not visible. \\
\addlinespace
Narrative \textit{N} & Description necessarily containing the body part influencing the emotion. Involves an emotional undertone and scene setting. \\
\addlinespace
Body Parts \textit{B} & Identifiable body parts involved in emotion expression. \\
\bottomrule
\end{tabular}
}
\caption{Components of structured textual output from LVLMs.} 

\label{tab:output_format}
\end{table}

\begin{figure}[h]
\centering
  \includegraphics[width=0.99\columnwidth]{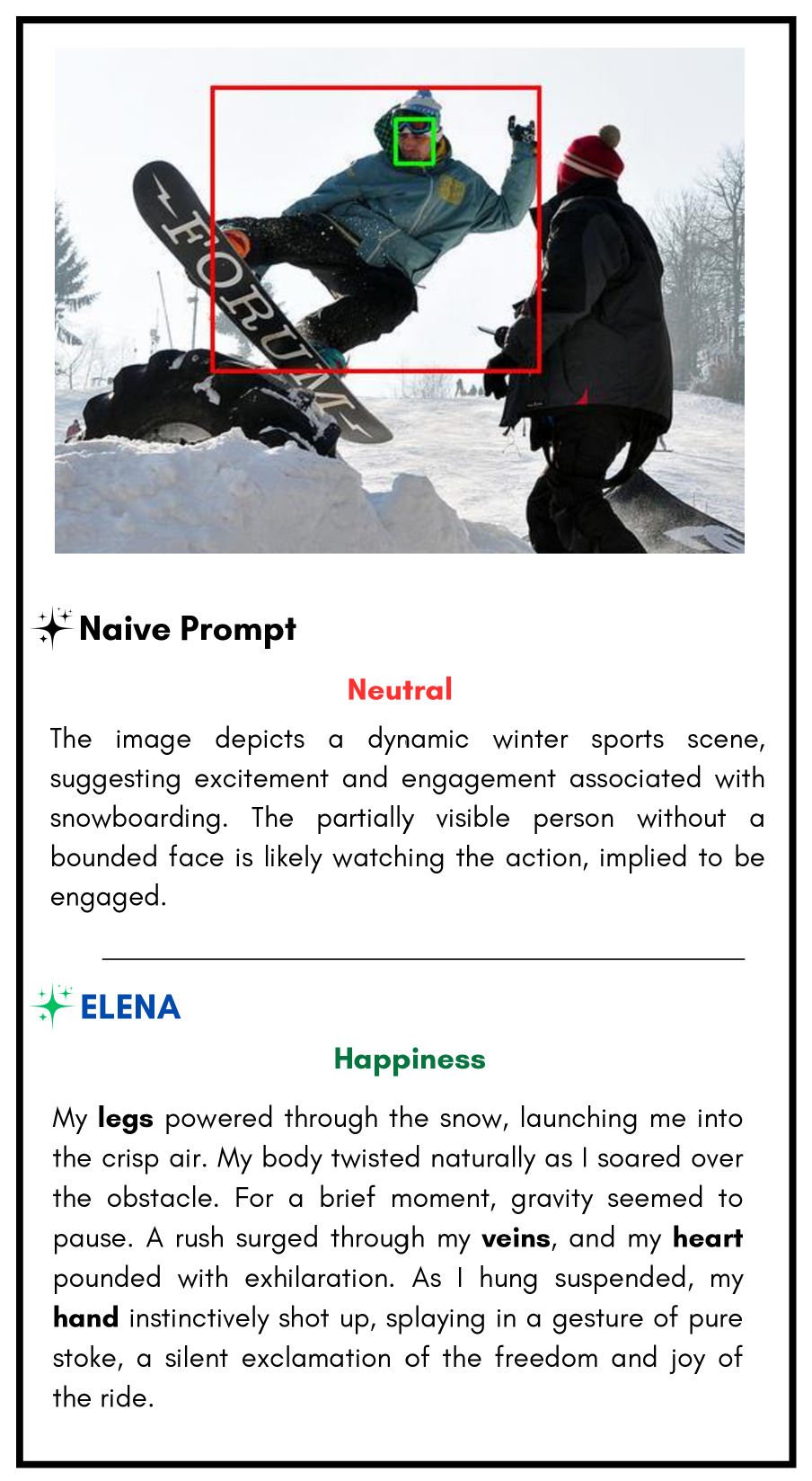}
  \caption{Example of ELENA compared to generic emotion response. Green highlights correctly predicted labels, while red indicates an incorrect prediction. The red box represents the coordinates of the person's body; the green bounding box is generated using YuNet. The sample is from the EMOTIC dataset. }
  \label{fig:example}
\end{figure}

We evaluated a suite of proprietary and open-source models, notably Gemini 2.5 Flash \citep{geminiteam2025geminifamilyhighlycapable} and Gemma-3-12B \citep{gemmateam2025gemma3technicalreport} and Llama-3.2-11B/90B  \citep{grattafiori2024llama3herdmodels} respectively. Due to computational constraints, we only tested Llama's 90B version on the smaller dataset (BESST, Appendix~\ref{sec:besst}).
Each narrative follows the definitions in Table \ref{tab:output_format}, with models free to use varied vocabulary within the structural constraints.

For each image, we also query the model to output one of the six corresponding Ekman emotions that match the embodied emotion displayed. If the image does not elicit a clear display of emotion, the model will predict the \textit{Neutral} label. For narratives, more emphasis was placed on describing body-related expressions without explicitly mentioning the emotion to steer the model away from simply restating the emotion. Figure~\ref{fig:example} shows an example response from the ELENA compared to a naive prompt. While the generic prompt captures the broad context of the image, it fails to isolate the fine-grained embodied expressions and lacks specificity in its description.

In the \textit{narrative} response, for the context, we use words like ``scene'' or ``story'' to encourage a portrayal-style explanation, allowing the model to come up with the reasoning as to what event and bodily reaction occurred that made the person depict the emotion expressed in the images. Finally, we explicitly request a list of physical features or body parts; this ensures that even if the descriptions and narrative already contain them, the model isolates them for downstream analysis. Doing this helps us in reliably extracting which body parts the model focuses on for extended statistical comparisons.
As a final note, ELENA employs a single, unified prompt that generates emotion labels, descriptions, and narratives simultaneously in one forward pass, ensuring a coherent and jointly reasoned output.
The general template for the prompts is available in Appendix \ref{sec:adddatasets}.
\subsection{Face Masking}

It is essential to consider other bodily expressions than the facial region as stimuli for emotion recognition, which is also the central belief of this paper. Furthermore, in many real-world situations, facial expressions may not be fully visible or recognizable, making it challenging to interpret emotions.  Therefore, we implement a face masking approach to evaluate whether LVLMs can effectively identify emotions from bodily expressions and to assess the robustness of our pipeline.

The masking procedure is formalized as: 
$$I_{m}(x,y) = 
\begin{cases} 
M, & \text{if } (x,y) \in \bigcup_{f \in F} R_f \\ 
I_{o}(x,y), & \text{otherwise} 
\label{eq:face_masking}

\end{cases}$$

The above equation depicts our facial masking procedure, where $I_{m}(x,y)$ represents the pixel value at coordinates $(x,y)$ in the masked image, $I_{o}(x,y)$ represents the original pixel value before masking, and $M$ is the mask color. The set $F$ contains all faces detected by YuNet \cite{wu2023yunet}, while $R_f$ denotes the region of the facial area. For more details on the masking framework, refer to Appendix~\ref{sec:facemasking}.
We opted for complete rather than partial face masking to create a more rigorous test of embodied emotion recognition, ensuring models cannot rely on any residual facial cues. Furthermore, partial masking of specific features, such as the eyes or mouth, is often unreliable in datasets like HECO and EMOTIC, which feature many non-frontal views and distant shots.

\subsection{Attention Visualization}

To investigate which regions of input images the LVLMs prioritize when processing emotion-related queries, we conducted an attention analysis using Llama-3.2-11B-Vision. We selected this model due to its open-weight architecture, computational efficiency, and widespread adoption in the research community, which promotes reproducibility. We used a bare-bone prompt (``What is the emotion displayed in the image?'') and ran a forward pass through the model to extract attention patterns from the cross-attention layers. We then computed attention heatmaps by averaging attention scores across all tokens within each cross-attention layer. To ensure computational tractability and avoid image tiling artifacts, we resized all images to the encoder's default resolution of 560×560 pixels. From each layer, we extract the first 1,600 visual tokens corresponding to the first image tile, where each token represented a 14×14 pixel patch of the input image. We then generated layer-specific attention heatmaps by computing the average attention score across all attention heads. These scores are finally mapped back to their corresponding spatial locations in the original image to create interpretable visualizations of the model's attention patterns.

\begin{table*}[t]
\footnotesize
\centering
\resizebox{0.95\linewidth}{!}{
\begin{tabular}{l l ccc ccc}
\toprule
\multirow{2}{*}{\textbf{Dataset}} & \multirow{2}{*}{\textbf{Model}} & \multicolumn{3}{c}{\textbf{Naive Prompt}} & \multicolumn{3}{c}{\textbf{ELENA}} \\
\cmidrule(lr){3-5} \cmidrule(lr){6-8}
 & & \textbf{Precision} & \textbf{Recall} & \textbf{F1} & \textbf{Precision} & \textbf{Recall} & \textbf{F1} \\
\midrule
\multirow{3}{*}{$\textbf{HECO}_\textit{Normal}$} & Gemini 2.5 Flash & 38.4 & 30.8 & 30.6 & 45.8 & 31.7 & 34.5$_{\textcolor{blue}{\uparrow3.9}}$ \\
 & Llama-3.2-11B & 33.2 & 26.6 & 26.2 & 37.1 & 29.3 & 29.5$_{\textcolor{blue}{\uparrow3.3}}$ \\
 & Gemma-3-12B & 37.0 & 31.7 & 30.6 & 38.4 & 30.1 & 30.1$_{\textcolor{red}{\downarrow0.5}}$ \\
\midrule
\multirow{3}{*}{$\textbf{HECO}_\textit{Masked}$} & Gemini 2.5 Flash & 27.4 & 19.2 & 16.9 & 33.6 & 34.7 & 31.5$_{\textcolor{blue}{\uparrow14.6}}$ \\
 & Llama-3.2-11B & 26.5 & 17.7 & 14.4 & 30.9 & 24.9 & 22.5$_{\textcolor{blue}{\uparrow8.1}}$ \\
 & Gemma-3-12B & 29.2 & 18.2 & 14.3 & 32.1 & 26.2 & 23.6 $_{\textcolor{blue}{\uparrow9.3}}$ \\
\midrule
\multirow{3}{*}{$\textbf{EMOTIC}_\textit{Normal}$} & Gemini 2.5 Flash & 43.9 & 25.1 & 28.6 & 42.8 & 22.6 & 27.7$_{\textcolor{red}{\downarrow0.9}}$ \\
 & Llama-3.2-11B & 44.2 & 20.0 & 23.3 & 33.0 & 21.0 & 21.5$_{\textcolor{red}{\downarrow1.8}}$ \\
 & Gemma-3-12B & 45.7 & 23.4 & 26.5 & 41.3 & 22.2 & 28.0 $_{\textcolor{blue}{\uparrow1.5}}$\\
\midrule
\multirow{3}{*}{$\textbf{EMOTIC}_\textit{Masked}$} & Gemini 2.5 Flash & 28.0 & 15.9 & 15.6 & 41.1 & 20.2 & 26.0$_{\textcolor{blue}{\uparrow10.4}}$ \\
 & Llama-3.2-11B & 35.9 & 16.1 & 17.1 & 32.6 & 17.5 & 20.7$_{\textcolor{blue}{\uparrow3.6}}$ \\
 & Gemma-3-12B & 39.3 & 17.8 & 17.2 & 39.1 & 19.7 & 24.8$_{\textcolor{blue}{\uparrow7.6}}$ \\
\midrule
\multirow{3}{*}{$\textbf{BESST}_\textit{Masked}$} & Gemini 2.5 Flash & 64.2 & 55.7 & 51.9 & 64.8 & 58.0 & 52.7$_{\textcolor{blue}{\uparrow0.8}}$ \\
 & Llama-3.2-11B & 36.8 & 30.2 & 22.2 & 48.3 & 44.9 & 37.8$_{\textcolor{blue}{\uparrow15.6}}$ \\
 & Gemma-3-12B & 63.6 & 39.0 & 32.3 & 55.5 & 46.9 & 41.6$_{\textcolor{blue}{\uparrow9.3}}$ \\
\bottomrule
\end{tabular}}
\caption{Performance comparison across datasets and prompt types. Values are presented as macro-averaged (\%). Subscripts denote image types: $_\textit{Normal}$ for unmasked images and $_\textit{Masked}$ for images with faces masked. $_{\textcolor{blue}{\uparrow}}$ indicates absolute F1 performance increase with EE prompt compared to Naive prompt, while $_{\textcolor{red}{\downarrow}}$ indicates absolute decrease.}
\label{tab:performance_comparison}
\end{table*}

\section{Results and Analysis}

With our methodology in place for eliciting embodied emotional indicators from LVLMs, we move to the experimental analysis and results. We evaluate ELENA's performance through two key comparisons: first, we assess how our framework performs against baseline naive prompts across different models and datasets, and second, we examine how face masking affects the model behavior and attention patterns in embodied emotion recognition tasks. We conclude with attention visualization analysis, breakdowns of emotion-specific performance and comparison to a modular baseline.

\subsection{ELENA vs. Naive Prompt Performance}

We compare ELENA's structured prompting approach against a baseline naive prompt to assess whether explicitly incorporating embodied emotion definitions enhances model recognition capabilities. Table~\ref{tab:performance_comparison} presents the performance results across three datasets for multiple LVLMs. ELENA consistently outperforms naive prompts in twelve out of fifteen experimental configurations, with the remaining three showing only narrow performance gaps. Among the evaluated models, Gemini 2.5 Flash achieves the highest F1 scores across most datasets and conditions, demonstrating robust overall performance. The Gemma and Llama models follow closely, yielding comparable results.
 Further analysis on the BESST dataset (Appendix~\ref{sec:besst}) confirms that Gemini 2.5 Flash surpasses even the larger Llama-3.2-90B model. 
 
The performance improvements vary depending on the characteristics of the dataset. On HECO, ELENA shows consistent gains ranging from 3.3 to 14.6 F1 points depending on the model and masking condition. For EMOTIC, results are more mixed, with slight performance drops in some unmasked scenarios, which we attribute to the dataset's rich contextual cues and multiple-person scenarios that can interfere with focused embodied emotion analysis. On BESST, improvements are substantial, with ELENA providing up to 15.6 F1 point gains.
These results demonstrate that structured prompting effectively guides LVLMs toward embodied emotion recognition by directing attention to bodily expressions rather than relying solely on facial features or contextual cues. The framework's effectiveness varies with dataset quality and image complexity, but it consistently provides meaningful improvements over generic emotion recognition approaches.

\noindent \textbf{Takeaway:} ELENA consistently outperforms naive emotion recognition prompts across multiple models and datasets, with performance gains ranging from modest (3-4 F1 points) to substantial (15+ F1 points).

\subsection{Impact of Face Masking on Embodied Emotion Recognition}

To evaluate ELENA's ability to recognize emotions from bodily expressions independent of facial indicators, we implemented a systematic face masking technique using the YuNet detection model, followed by complete occlusion of facial regions. This analysis reveals how models adapt when they rely exclusively on embodied indicators. We essentially duplicate the experiment for each model, with the only change being that the input image now has the face region obscured. We ensure that the prompt remains the same, except that it specifies not to focus on the facial area or the masked region.

We report that performance improvements under masking conditions are most pronounced, with ELENA achieving up to 15.6 F1-point gains over naive prompts (indicated by blue up arrows in the masked sections in Table~\ref{tab:performance_comparison}). This suggests that while LVLMs struggle with autonomous attention redirection when faces are masked, structured guidance can effectively compensate for this limitation.
From analyzing the performance on masked images, we identify two key insights. First, ELENA succeeds in shifting focus to mention body parts apart from the facial region. Table~\ref{tab:bodydistribution} depicts a substantial shift in highlighting body parts from various anatomical areas of the body. The model's description shifts to focus on the body's limbs, with a primary concentration on the hands, arms, shoulders, and legs. It also sees a percentage increase in the body's torso. Although not in the top ten, narratives still mention facial parts, suggesting that models are biased in concentrating on facial features. Other reasons, discovered through manual inspection, included issues with the YuNet model, specifically the failure to mask faces when a scene is filled with a crowd with averted bodies. Another observation is that with the faces masked, models lean towards more fine-grained details surrounding the context.\newline
Second, multiple images are not left with ample context after masking. This can be directly attributed to the images in the dataset, particularly those from the EMOTIC and HECO datasets. Images might be captured in such a way that the facial region is the center of interpretation for the model. Other body parts were rarely visible for the ELENA framework to be effective. Once the face is masked, the image produces very ambiguous and non-coherent narratives, which in turn lead to incorrect label predictions. This highlights a major issue in prevalent emotion-recognition image datasets and further underscores the need for a new, qualitatively filtered dataset. Further arguments on this result are discussed in Appendix \ref{sec:addres}.

\begin{table}[t]
\centering
\resizebox{0.98\linewidth}{!}{
\begin{tabular}{lr|lr}
\toprule
\textbf{Normal Images} & \textbf{\%} & \textbf{Face-Masked Images} & \textbf{\%} \\
\midrule
eye & 14.45 & \textbf{hand} & \textbf{21.82} \\
{mouth} & 9.66 & \textbf{arm} & \textbf{18.62} \\
heart & 9.29 & shoulder & 11.76 \\
hand & 9.15 & heart & 5.87 \\
chest & 5.60 & chest & 4.79 \\
arm & 3.95 & torso & 4.49 \\
shoulder & 3.92 & leg & 4.24 \\
{face} & 3.36 & finger & 4.16 \\
{lip} & 2.98 & body & 3.60 \\
head & 2.98 & head & 1.80 \\
\midrule
\multicolumn{4}{c}{\textbf{Anatomical Regions}} \\
\midrule
{Head/Face} & {40.63} & {\textbf{Limbs}} & {\textbf{50.18}} \\
{Limbs} & {17.14} & Torso & 22.98 \\
Internal/Conceptual & 13.88 & Other & 14.07 \\
Torso & 13.21 & Internal/Conceptual & 8.05 \\
Other & 15.13 & \textbf{{Head/Face}} & \textbf{{4.71}} \\
\bottomrule
\end{tabular}
}
\caption{ Top ten body part mentions (\%) by ELENA on face masked versus unmasked images. The lower section categorizes all mentions of body parts. The distribution is based on the HECO image dataset.  \textbf{Bold} highlights dramatic shifts.}
\label{tab:bodydistribution}
\end{table}

\noindent \textbf{Takeaway:} Face masking reveals that ELENA successfully redirects model attention from facial features to alternative body parts, thus achieving better performance than baselines.

\subsection{Attention Map Analysis}

\begin{figure}[t]
    \centering
    \includegraphics[width=0.9\linewidth]{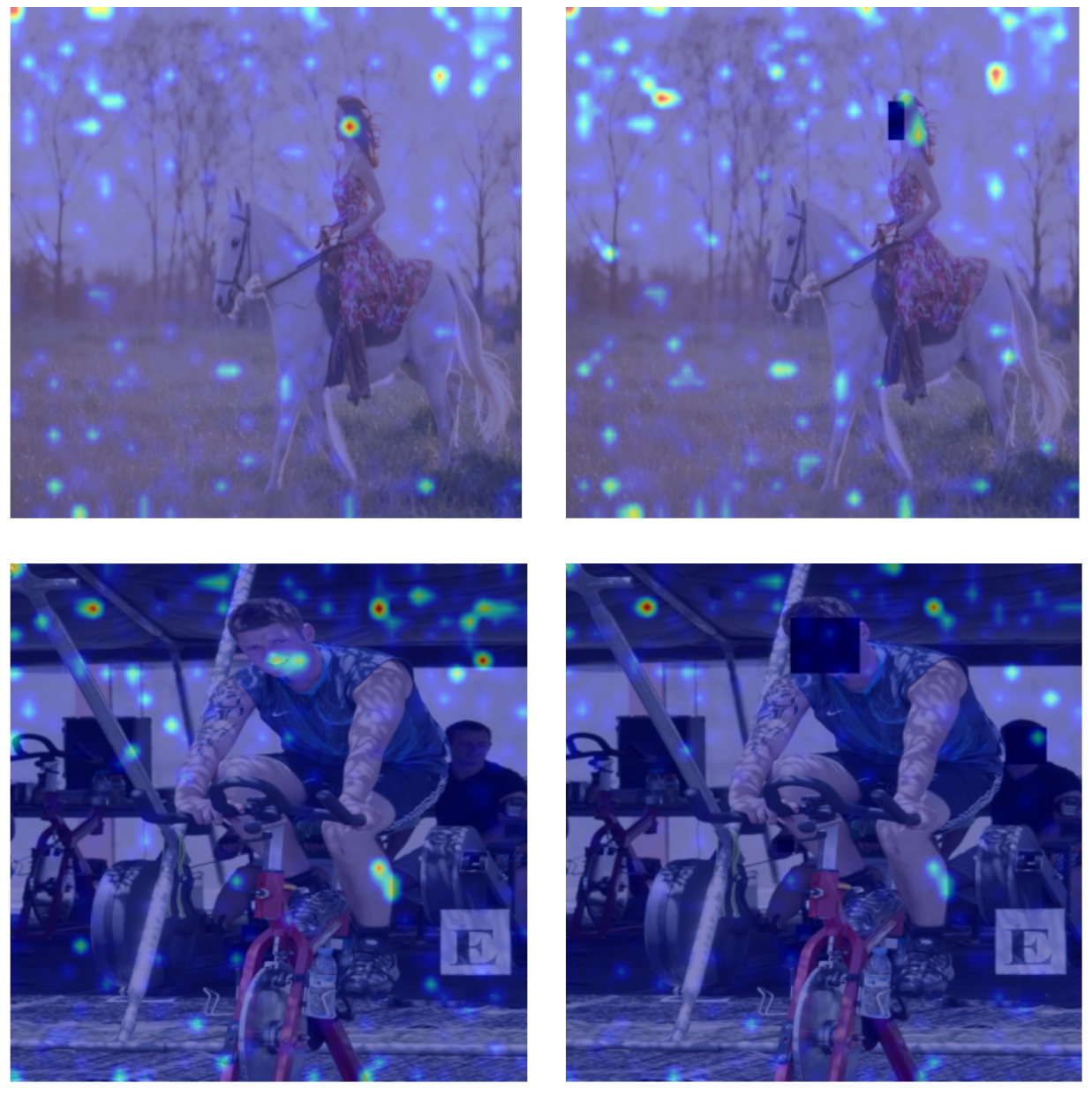}
    \caption{Visualization of Llama-3.2-11B's cross-attention layer 13 for HECO images.}
    \label{fig:attn-viz}
\end{figure}

\begin{figure}[t]
  \includegraphics[width=\columnwidth]{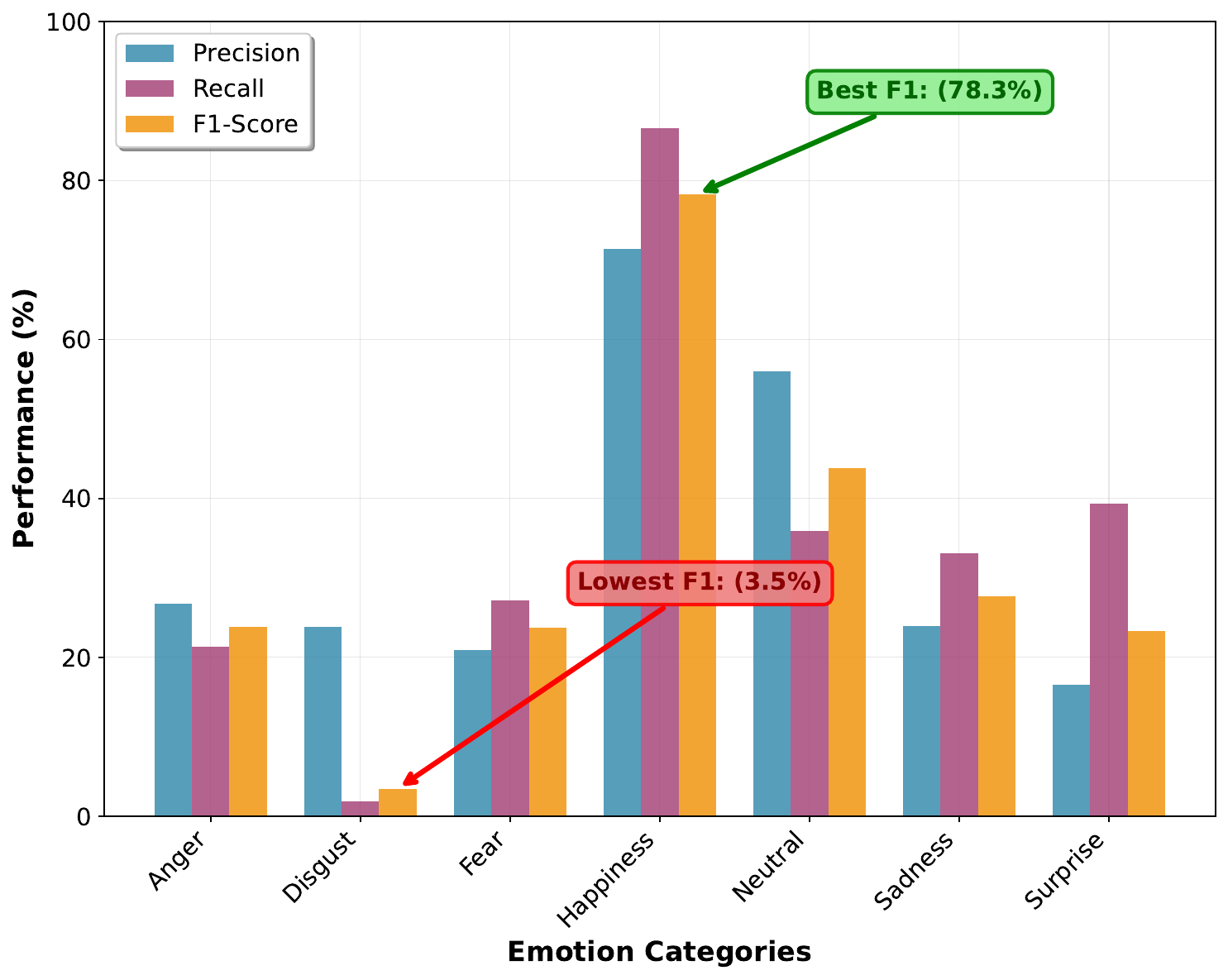}
  \caption{Per-Category Analysis of each Emotion Label. Results are from the Gemini-2.5-Flash Model on the HECO dataset.}
  \label{fig:per-metrics}
\end{figure}

\begin{figure}[t]
    \centering
    \includegraphics[width=0.99\linewidth]{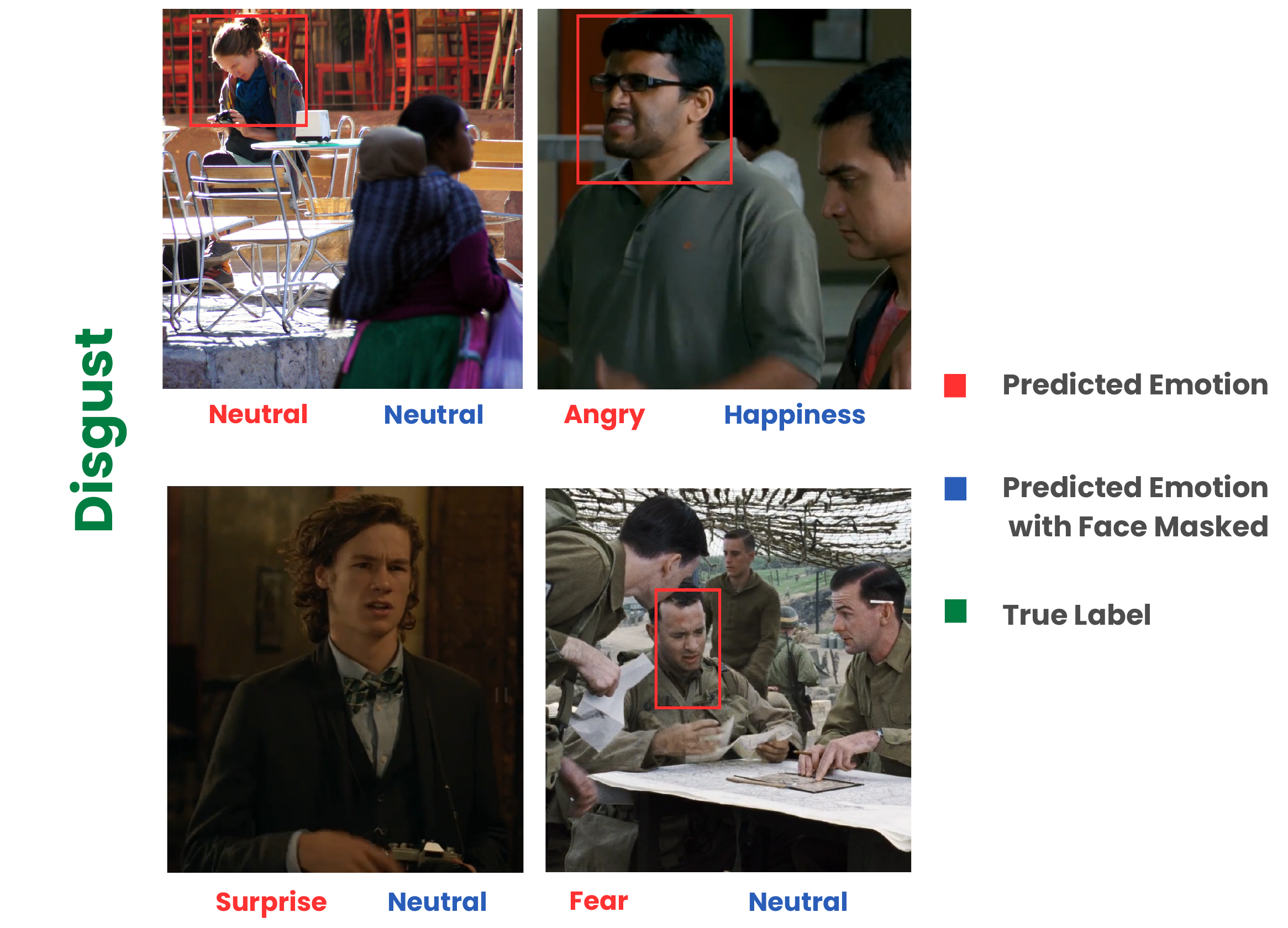}
    \caption{Failure cases of predicting the label \textit{Disgust} in both unmasked and masked conditions. Examples are from the HECO dataset.
    }
    \label{fig:failurecase}
\end{figure}

\begin{figure}[t]
    \centering
    \includegraphics[width=0.99\linewidth]{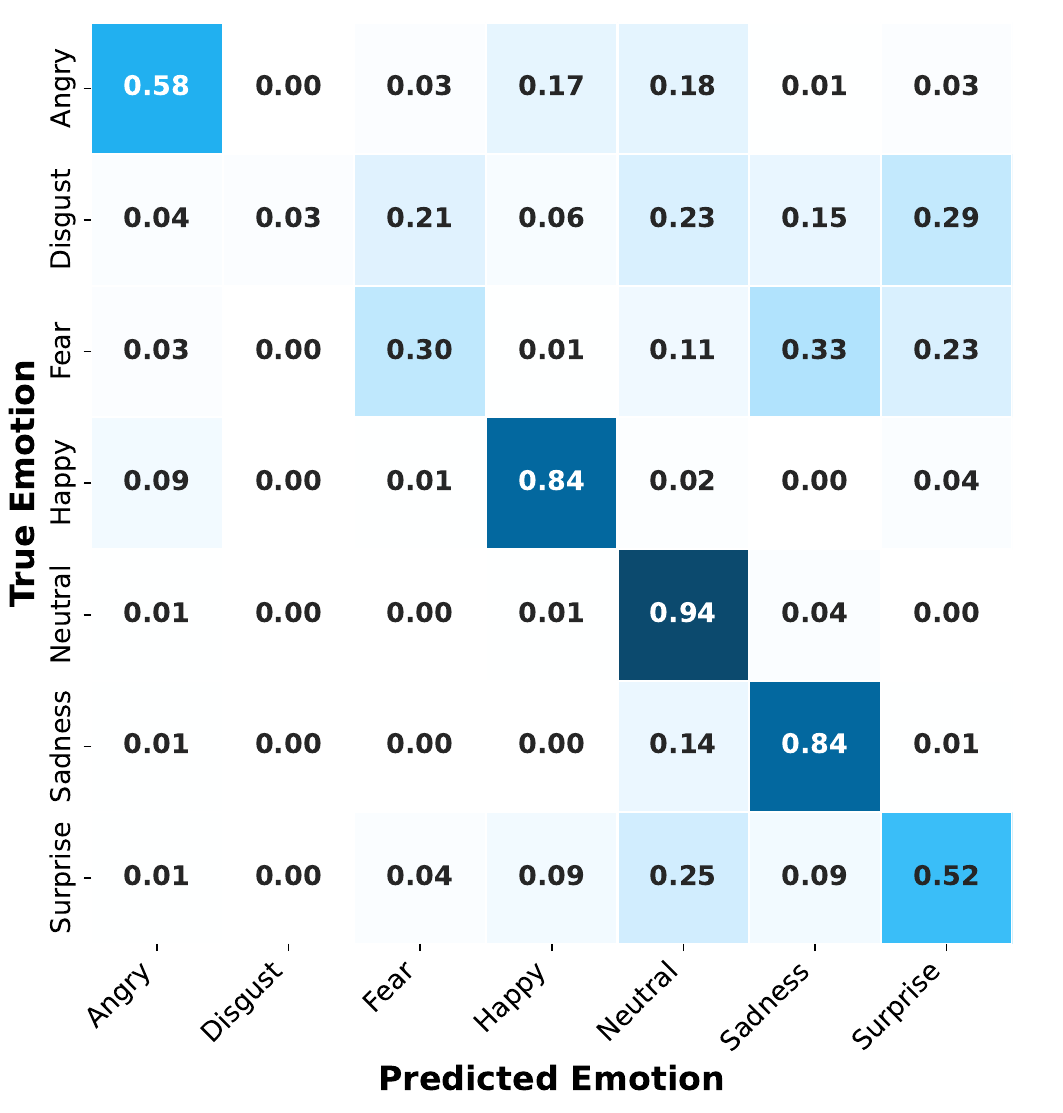}
    \caption{Heatmap visualization of true versus predicted labels of the BESST dataset. Results are from Gemini-2.5-Flash.
    }
    \label{fig:heatmap-viz}
\end{figure}

Through analysis of the attention maps obtained, we identified distinct patterns in Llama-3.2-11B-Vision's behavior during emotion recognition tasks. The model exhibits a pronounced bias toward facial features, particularly the mouth region, in cross-attention layers 13-28 when processing unmasked images. When presented with masked faces, the model demonstrates two problematic response patterns, neither of which constitutes effective adaptation. As illustrated in Figure \ref{fig:attn-viz}, the model either continues to direct attention toward the masked facial region and its immediate surroundings, essentially attempting to extract information from areas that no longer provide meaningful emotional cues, or it diverts attention away from the face entirely without increasing focus on alternative emotional indicators such as body language. Crucially, the model's attention to non-facial regions remains at baseline levels, comparable to those in unmasked scenarios, indicating a failure to recognize and compensate for the loss of facial information. This behavior could explain the performance degradation observed in masked images, as shown in Table \ref{tab:performance_comparison}.

The findings indicate a fundamental bias in the employed model's attention mechanism: the model does not redirect its focus to non-facial emotional indicators when primary facial indicators are unavailable. Rather than attending to alternative information sources such as body posture or gesture, the model exhibits erratic attention patterns that compromise its effectiveness in masked scenarios. This limitation warrants further investigation, as emotion is fundamentally grounded in the whole body through embodied emotion, extending beyond facial expressions alone. Models that rely predominantly on facial features may demonstrate only a superficial understanding of human emotional expression, suggesting a critical gap in current vision-language model architectures that future research should address to achieve more robust emotion recognition capabilities.

\noindent \textbf{Takeaway}: The employed LVLM (Llama-3.2-11B) exhibits fundamental attention adaptation failure, maintaining facial bias even when faces are masked, and failing to compensate by increasing focus on informative body regions, revealing an architectural limitation that explains the performance degradation in embodied emotion recognition.

\subsection{Emotion-Specific Performance Analysis}

Figure~\ref{fig:per-metrics} shows the model's performance for each emotion category and identifies substantial disparities between them. \textit{Happiness} achieves near-ceiling scores, reaching approximately 90\% for the best model on the HECO dataset. This high accuracy can be attributed primarily to the distinctive visual cues associated with \textit{Happiness}---particularly smiling faces and highly expressive body language.

In contrast, the recognition of \textit{Disgust} remains notably problematic, with frequent misclassifications. An example is shown in Figure \ref{fig:failurecase}. Most of the labels get majorly misclassified into three other labels: \textit{Fear, Sadness, and Surprise}, along with the \textit{Neutral} tag, as evident in Figure \ref{fig:heatmap-viz}. The difficulty arises due to definitional ambiguity and subtlety in visual representation, making disgust inherently challenging for models to identify reliably. Another occurrence we came across was the model's refusal to provide an answer to the input query where the gold annotation for the image was the \textit{Disgust} label. We can attribute this partially to the nature of the image; however, it could also be due to the sensitivity of the emotion, which triggers the safety measures of LVLMs. Therefore, this issue is more likely due to the strictness of content filtering rather than a problem with the model's understanding or inappropriate images. Furthermore, this emotion is significantly underrepresented in the datasets, which exacerbates the difficulty in classification by providing insufficient training examples for equitable learning.

Additionally, face masking notably increases the prediction frequency of the \textit{Neutral} category. This phenomenon occurs when substantial information loss results from the obscuring of informative facial expressions, leaving the models dependent solely on body posture cues. Given that body postures are typically less distinctive and more ambiguous, the models default to predicting \textit{Neutral} in uncertain cases, thus reducing recall for expressive emotional categories. The predicted emotion labels are also scrutinized using a dimensional valence-arousal-dominance (VAD) model. The analysis regarding the VAD scores is shown in Appendix \ref{sec:appendix_vad}.

\noindent \textbf{Takeaway}: Our analysis reveals significant performance disparities across emotions; models excel at identifying specific labels, e.g., \textit{Happiness}, but consistently misclassify others, notably, \textit{Disgust}, due to its visual subtlety in the bodily cues and the nature of definitional ambiguity in images.

\subsection{Comparison with Modular Baselines}

To further contextualize ELENA's performance, we compare it with a supervised modular design. We first apply a widely-used image captioning model, LLaVA-1.5-7b \cite{liu2024improvedbaselinesvisualinstruction}, to get the caption of an image, then fine-tune a BERT-base model~\citep{devlin-etal-2019-bert} to predict the emotion label. Here we experiment on the \textbf{HECO} dataset. The choice of this dataset stems from our manual inspection, as it provides a better representation of 'emotion-centric' images in a natural setting. This experiment was designed to assess whether standard captioning can retain the subtle, context-rich bodily expressions central to our task. 

\begin{table}[t]
\centering
\small
\begin{tabular}{@{}lccc@{}}
\toprule
\textbf{Method} & \textbf{Precision} & \textbf{Recall} & \textbf{F1} \\
\midrule
LLaVA + Fine-tuned BERT & 22.4 & 21.5 & 20.8 \\
\textbf{ELENA}  & 45.8 & 31.7 & 34.5 \\
\bottomrule
\end{tabular}
\caption{Performance comparison of the image-captioning-plus-classifier (LLaVA + BERT) baseline to Gemini 2.5 Flash on the HECO dataset.}
\label{tab:heco_baseline}
\end{table}
Our framework shows substantial improvement compared to the LLaVA with a fine-tuned BERT model. The performance gap reveals that embodied emotion information is not sufficiently captured while captioning. For instance, while LLaVA might generate a generic caption like ``A person standing alone in a field, lost in thought'', ELENA's framework would further capture specific embodied indicators, such as ``shoulders slumped with head tilted down, suggesting dejection.'' This analysis reveals that standard image captions often fail to convey the subtle bodily expressions that are crucial to the embodied emotion task for affective experiences.

\section{Conclusion}

In this work, we introduced ELENA, a framework that employs LVLMs in a zero-shot setting to generate structured narratives that anatomize embodied emotions. Our experiments reveal that ELENA yields credible improvements over naive prompts, demonstrating the models' ability to adapt in both unmasked and masked settings despite information loss. Our experiments showed that the Gemini 2.5 Flash model generally outperformed other models across most datasets. However, labels like \textit{Disgust} proved particularly challenging to identify from bodily expressions. The attention analysis provides a diagnosis by documenting persistent facial bias. It establishes an essential empirical foundation that enables the community to develop targeted solutions, such as fine-tuning with redirected attention mechanisms, for accurate anatomical understanding of embodied emotions.

\section*{Limitations}

The proposed work recognizes some of the barriers it encounters in its formation. First, masking the face leaves little to no information in the image, primarily due to the nature of the image. In such cases, the model is forced to output \textit{Neutral} as a response, and this issue extends to the current image datasets, which may not elicit a clear emotional reaction from the person(s) involved, subsequently leading to incorrect analysis, even in human annotations. Second, a direct comparison between ELENA narratives and the CHEER dataset \citep{zhuang-etal-2024-heart, duong-etal-2025-cheer} would be valuable. However, differences in style make alignment difficult: ELENA outputs are longer and encompass multiple body parts, incorporating a significant amount of context, whereas CHEER sentences are more concise. Therefore, individual descriptions from ELENA need to be broken down into aspect categories to facilitate further analysis of their impact. Third, we believe ELENA-generated narratives could benefit downstream applications such as emotion recognition, conversational agents, or human-robot interaction. We view this as an important direction for future work.

\section*{Acknowledgments}

We thank the CincyNLP group for their constructive feedback. We also appreciate the anonymous EMNLP reviewers for their comments and suggestions, which helped us improve this paper.

\bibliography{custom}

\newpage
\appendix

\label{sec:appendix}

\section{VAD Score Interpretation}
\label{sec:appendix_vad}

Apart from narrative description and emotion labels, we also asked the model to provide us with the valence, arousal, and dominance scores. These scores are typically part of many emotion recognition analyses and, therefore, are further explored in detail. Figure 5 reveals distinctive patterns across the seven basic emotions. The \textit{Happiness} label emerges as the highest Valence score emotion (M = 8.51; where M is the mean value of this label), exhibiting substantially higher ratings than all other emotions, which cluster below the midpoint except for surprise (M = 5.22) and neutral states (M = 5.04). Regarding arousal, \textit{Fear}, \textit{Anger}, and \textit{Surprise} demonstrate the highest activation levels (M = 7.38 and 7.32, respectively), whereas neutral expressions predictably show reduced arousal (M = 3.51). This makes sense because the \textit{Neutral} label typically applies to images where there is no body enactment or emotion-intensive activity. In dominance scores, \textit{Happiness} maintains a relatively high dominance (M = 6.28), while \textit{Fear} exhibits markedly low dominance (M = 3.08), reflecting the characteristic vulnerability of the label.

\begin{figure}[h]

  \includegraphics[width=0.99\linewidth]{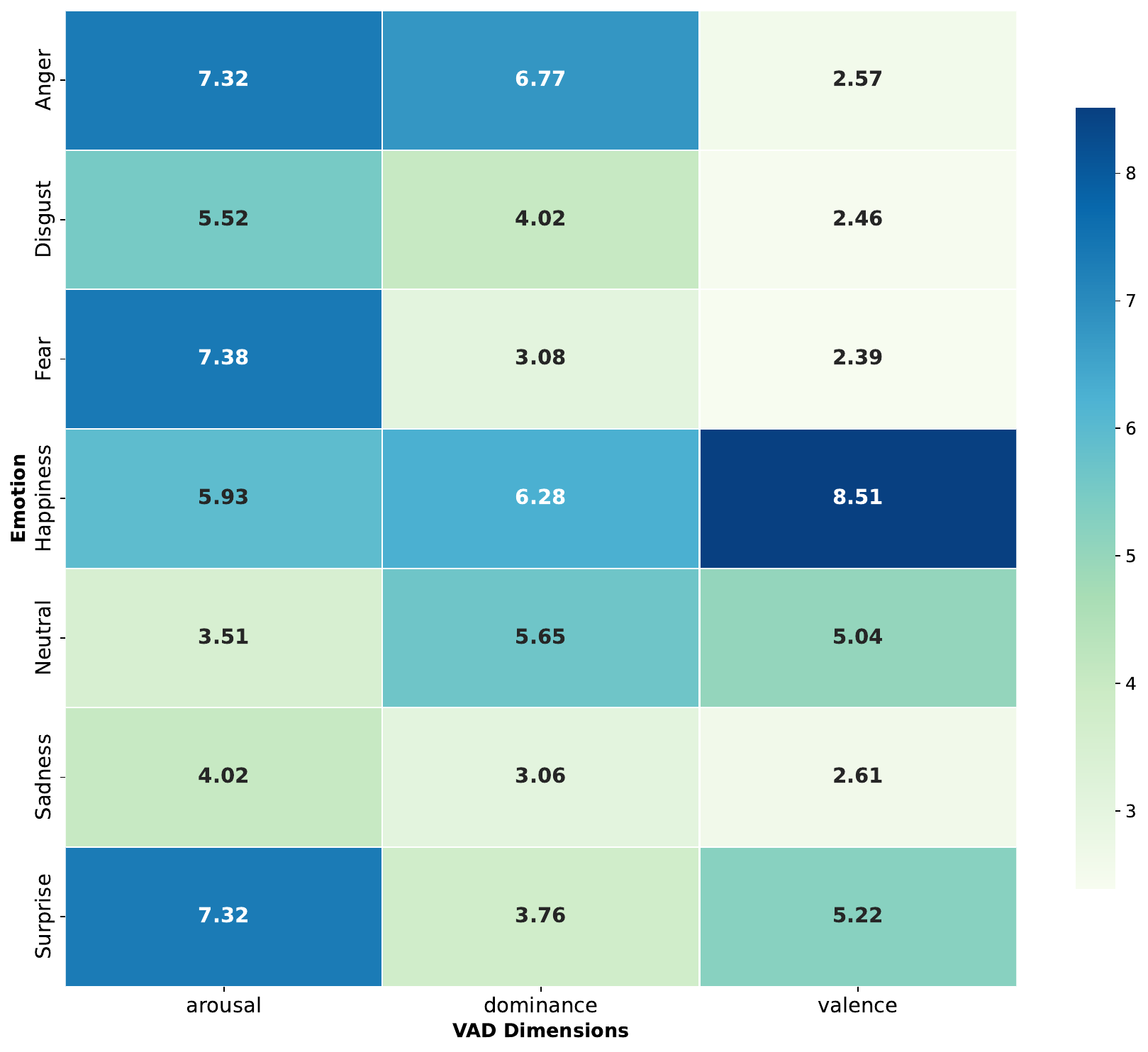}
  \caption {Valence, Arousal, and Dominance score representation associated with each emotion label. Results are from the HECO dataset.}
\end{figure}

\begin{figure}[t]

  \includegraphics[width=\columnwidth]{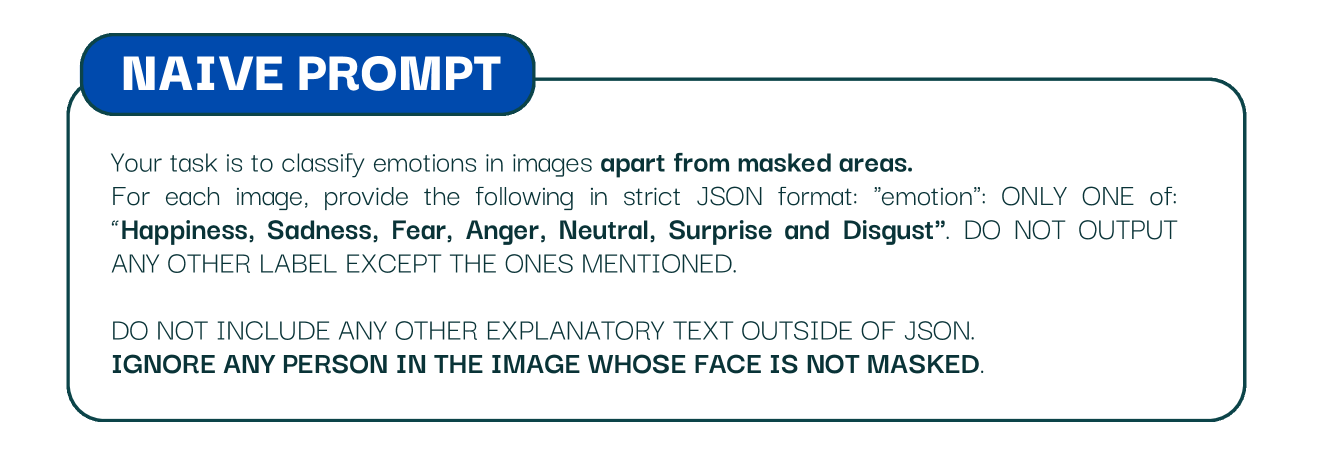}
  \caption {Naive prompt for masked images.}
  \label{fig:naive_prompt}
\end{figure}

\begin{figure}[t]

  \includegraphics[width=\columnwidth]{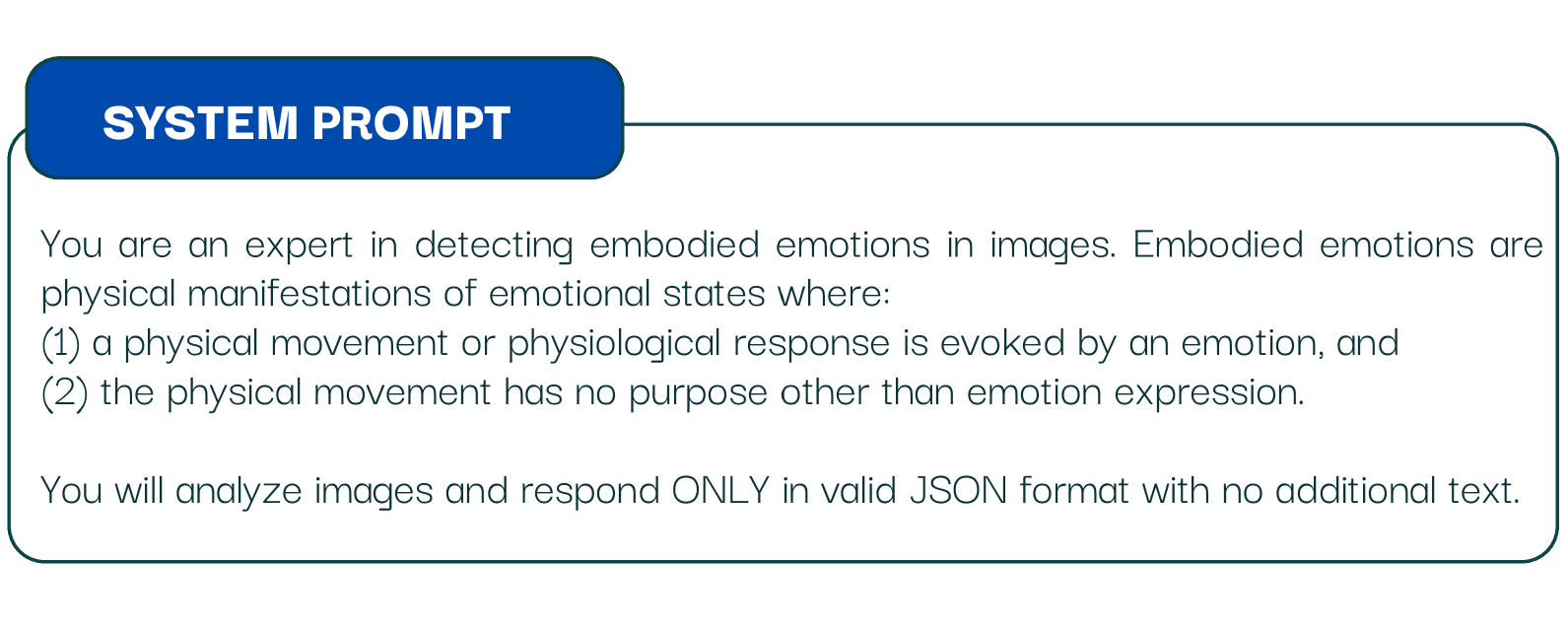}
  \caption {System prompt.}
  \label{fig:sys_prompt}
\end{figure}

\begin{figure}[t]

  \includegraphics[width=\columnwidth]{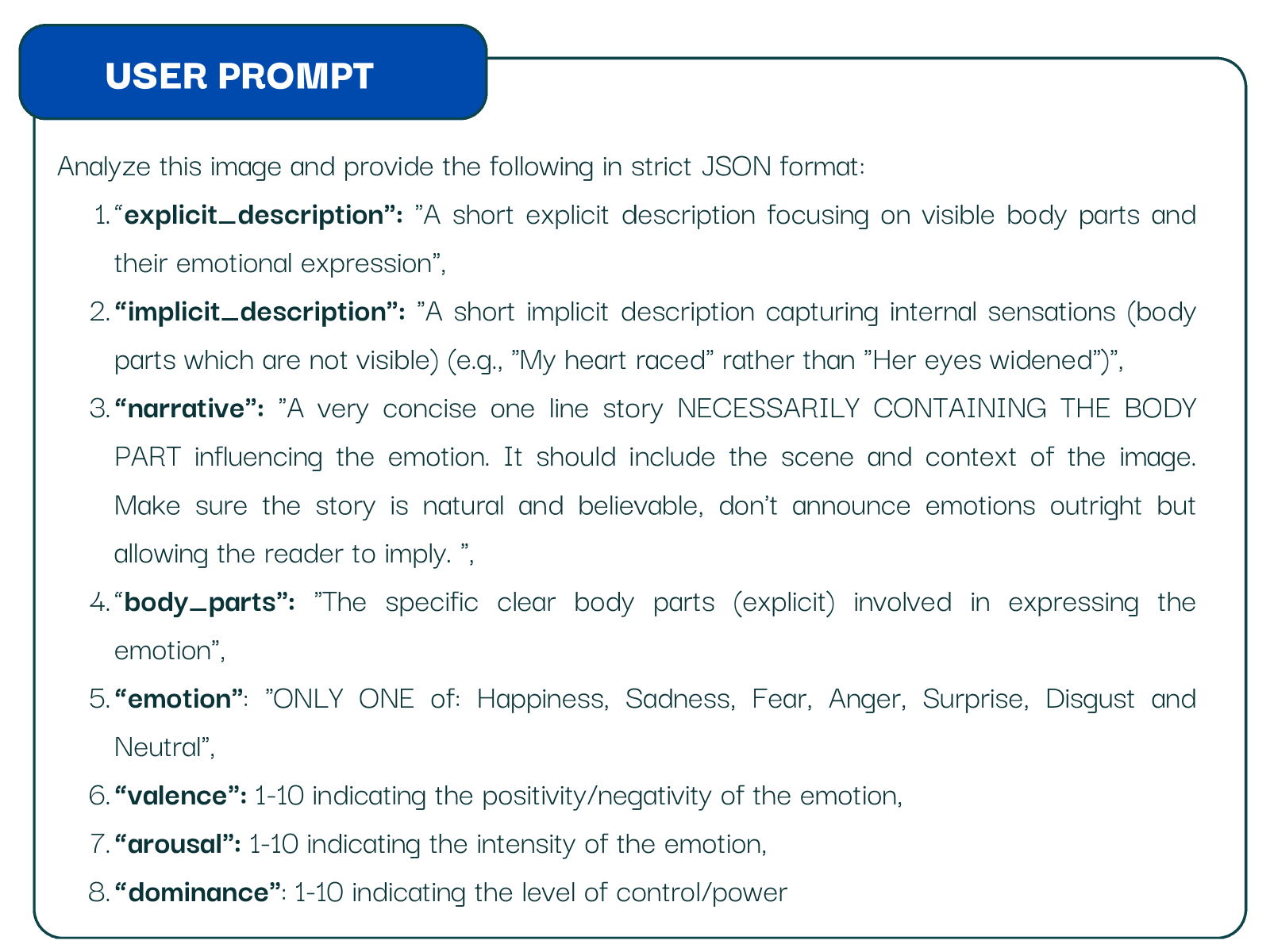}
  \caption {User prompt.}
  \label{fig:user_prompt}
\end{figure}

\section{Prompts Usage}
\label{sec:adddatasets}
We employed two prompting strategies in our experiments. First, a naive prompt (Figure~\ref{fig:naive_prompt}) asks the models to classify the emotion displayed in the image using a fixed label set, with minimal guidance and no multi-perspective interpretation. Importantly, it instructed the model to ignore any person whose face was not masked and to output only a single emotion label in JSON format. This setting reflects a constrained classification setup that lacks deeper narrative or reasoning components. In contrast, our primary prompt template employed a structured prompting scheme, combining a system message with a detailed user prompt (see Figures~\ref{fig:sys_prompt} and \ref{fig:user_prompt}). The system prompt defines the task as embodied emotion recognition and frames the model as an expert in this domain, emphasizing bodily movements as primary indicators of emotion.

The user prompt specifies the expected output format and fields, guiding the model to produce a multi-perspective emotional interpretation grounded in bodily expression. Responses were constrained to strict JSON format, enabling consistent parsing and analysis. All prompts were applied uniformly across images and models without post-hoc adjustment. This ensured comparability between outputs and minimized linguistic variance introduced by prompt phrasing. The prompt design plays a central role in eliciting structured, interpretable outputs from vision-language models in our study.

\begin{table}[t]
\centering
\begin{tabular}{|p{3.25cm}|p{3.25cm}|}
\hline
\textbf{EMOTIC Emotions} & \textbf{Ekman Emotion} \\
\hline
Affection & Happiness \\
Engagement & \\
Excitement & \\
Happiness & \\
Pleasure & \\
\hline
Disconnection & Sadness \\
Embarrassment & \\
Fatigue & \\
Pain & \\
Sadness & \\
Sensitivity & \\
Suffering & \\
Yearning & \\
\hline
Aversion & Disgust \\
Disapproval & \\
\hline
Disquietment & Fear \\
Fear & \\
\hline
Anger & Anger \\
Annoyance & \\
\hline
Anticipation & Neutral \\
Confidence & \\
Doubt/Confusion & \\
Esteem & \\
Peace & \\
\hline
Surprise & Surprise \\
\hline
\end{tabular}
\caption{Mapping of EMOTIC emotions to Ekman's six with a \textit{Neutral} category.}
\label{tab:emotion_mapping}
\end{table}

\subsection{Emotion Mapping and Reasoning Analysis}
\label{emomap}

We adopt the six basic Ekman emotions---Happiness, Sadness, Anger, Fear, Disgust, and Surprise—along with a Neutral category, as our primary label set. This choice strikes a balance between interpretability and coverage, enabling standardized evaluation while remaining expressive enough to capture a wide range of embodied affect.

This unified 7-category taxonomy was essential for enabling a consistent cross-dataset analysis, as our other datasets were natively annotated with similar categories. Consequently, a direct comparison using mean Average Precision (mAP) against prior EMOTIC benchmarks would be incongruous, as our framework evaluates a fundamentally different task grounded in embodied theory rather than multi-label classification on the 26 categories.

To justify this abstraction, we map each basic emotion to a broader set of fine-grained emotion categories from the EMOTIC dataset (Table~\ref{tab:emotion_mapping}). For example, the Sadness category includes not only sadness itself but also states such as fatigue, suffering, and yearning. Similarly, Happiness subsumes affective expressions like affection, engagement, and pleasure.

This mapping serves two key purposes: 
\begin{itemize}
    \item It enables aggregation of subtle and diverse affective cues under a shared emotional umbrella, facilitating more robust analysis of model predictions.
    \item It provides a reasoning structure for validating model outputs against grounded emotional expressions.
\end{itemize}

While models often use varied language to describe emotion, our mapping enables alignment between expressive counterparts and a more simplified label assignment.
\begin{figure*}[t]
\centering
\includegraphics[width=0.98\textwidth]{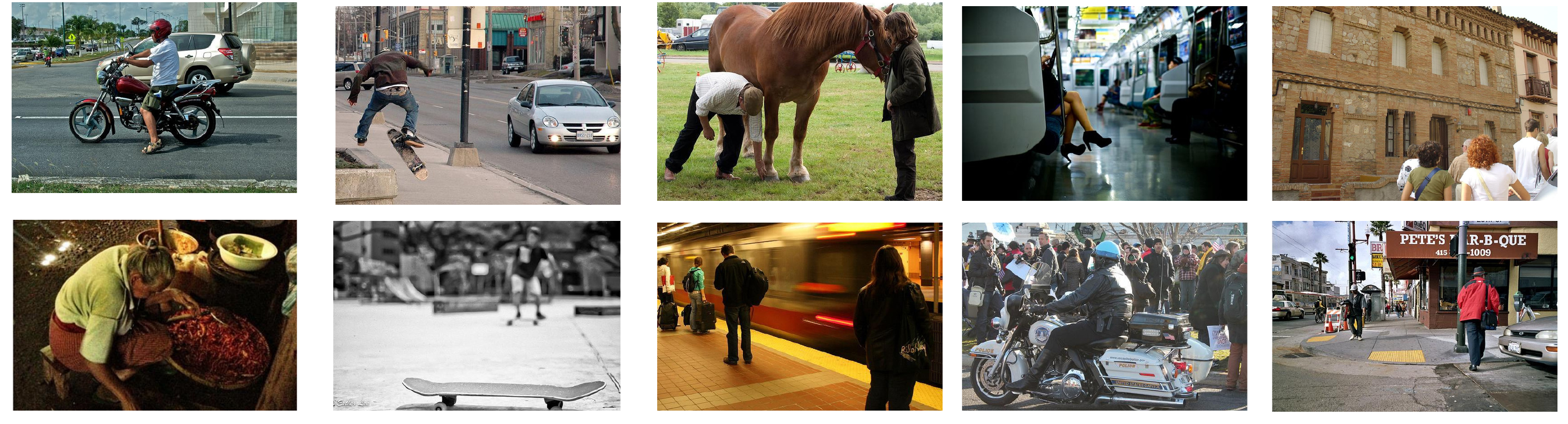}
\caption{Representative examples from the EMOTIC dataset illustrating challenges for embodied emotion recognition. Many images contain crowded places, distant subjects, ambiguous emotional contexts, or insufficient visibility of bodily expressions, making reliable analysis difficult, even with provided bounding box annotations.}
\label{fig:dataset_challenges}
\end{figure*}

\subsection{Dataset Quality Challenges for Embodied Emotion Recognition}
\label{dataqual}

While existing emotion recognition datasets, such as EMOTIC, provide bounding box annotations for people in images, many instances present significant challenges for embodied emotion analysis, as illustrated in Figure~\ref{fig:dataset_challenges}. Our manual analysis reveals that a substantial proportion of images in current emotion recognition datasets are inherently unsuitable for embodied emotion recognition tasks, and sometimes for general emotion recognition too, despite containing valid emotion annotations for general affective computing applications.

The fundamental issue stems from the conceptual difference between contextual emotion recognition and embodied emotion analysis. Embodied emotion recognition specifically requires clear, observable bodily expressions such as posture, gesture, limb positioning, and torso orientation. Many images in datasets contain subjects where these crucial embodied indicators are either absent, obscured, or ambiguous, making reliable analysis difficult even with sophisticated prompting frameworks like ELENA.

Consider the common scenario where multiple individuals appear in a single image, each with varying emotional states. While bounding box annotations attempt to isolate the target subject, the presence of other individuals with potentially conflicting emotional expressions creates visual noise that can mislead the LVLMs. This problem is particularly acute in crowded scenes or social gatherings where the emotional dynamics are complex and the target person's individual embodied expression becomes difficult to distinguish from the collective group behavior. Distance and viewing angle present additional complications. Many images capture subjects at distances where fine-grained bodily details necessary for embodied emotion recognition, such as hand positioning, shoulder tension, or subtle postural shifts, are not visible at sufficient resolution. Similarly, non-frontal views, while valuable for general emotion recognition, often obscure key body parts that serve as primary indicators in embodied emotion analysis. As our experiments demonstrate, these limitations become more pronounced when faces are masked, since the model must rely entirely on these often-inadequate bodily cues.

Perhaps most critically, we observe a systematic bias toward contextually inferred emotions rather than expressions grounded in observable bodily manifestations. For instance, an image might show a person labeled as \textit{Sad} based on the situational context (e.g., standing alone in the rain), yet their actual body posture may not exhibit the characteristic embodied markers of sadness. This disconnect between contextual emotion labels and embodied emotional expressions creates a fundamental training-evaluation mismatch for models designed to recognize emotions through bodily cues.

The implications extend beyond dataset quality to model evaluation. When a significant portion of the evaluation data contains ambiguous or unsuitable examples, performance metrics become unreliable indicators of the actual model's capability. Models may appear to perform poorly on embodied emotion recognition, not due to architectural limitations, but because the evaluation framework includes many instances where even human annotators would struggle to identify clear embodied emotional indicators. This challenge necessitates either more stringent data filtering protocols or the development of specialized datasets explicitly designed for recognizing embodied emotions.

\section{Interpreting LVLM Emotion Predictions on BESST}
\label{sec:besst}
\begin{figure}[t]
  \includegraphics[width=\columnwidth]{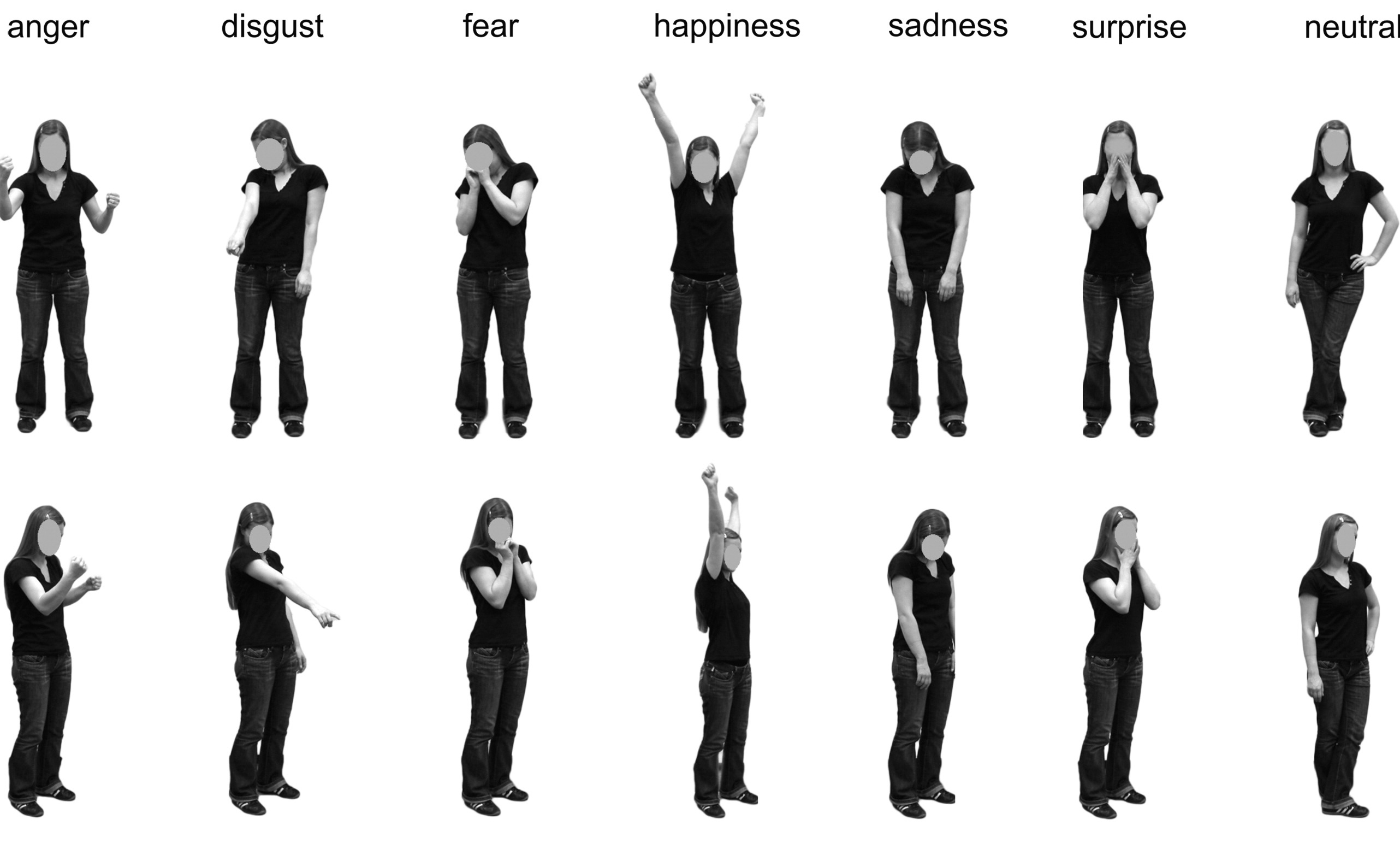}
  \caption{Example from the Bochum Emotional Stimulus Set (BESST), which depicts enactment of defined emotions in full frontal and averted view.}
  \label{fig:besst}
\end{figure}

\begin{figure}[t]
  \includegraphics[width=\columnwidth]{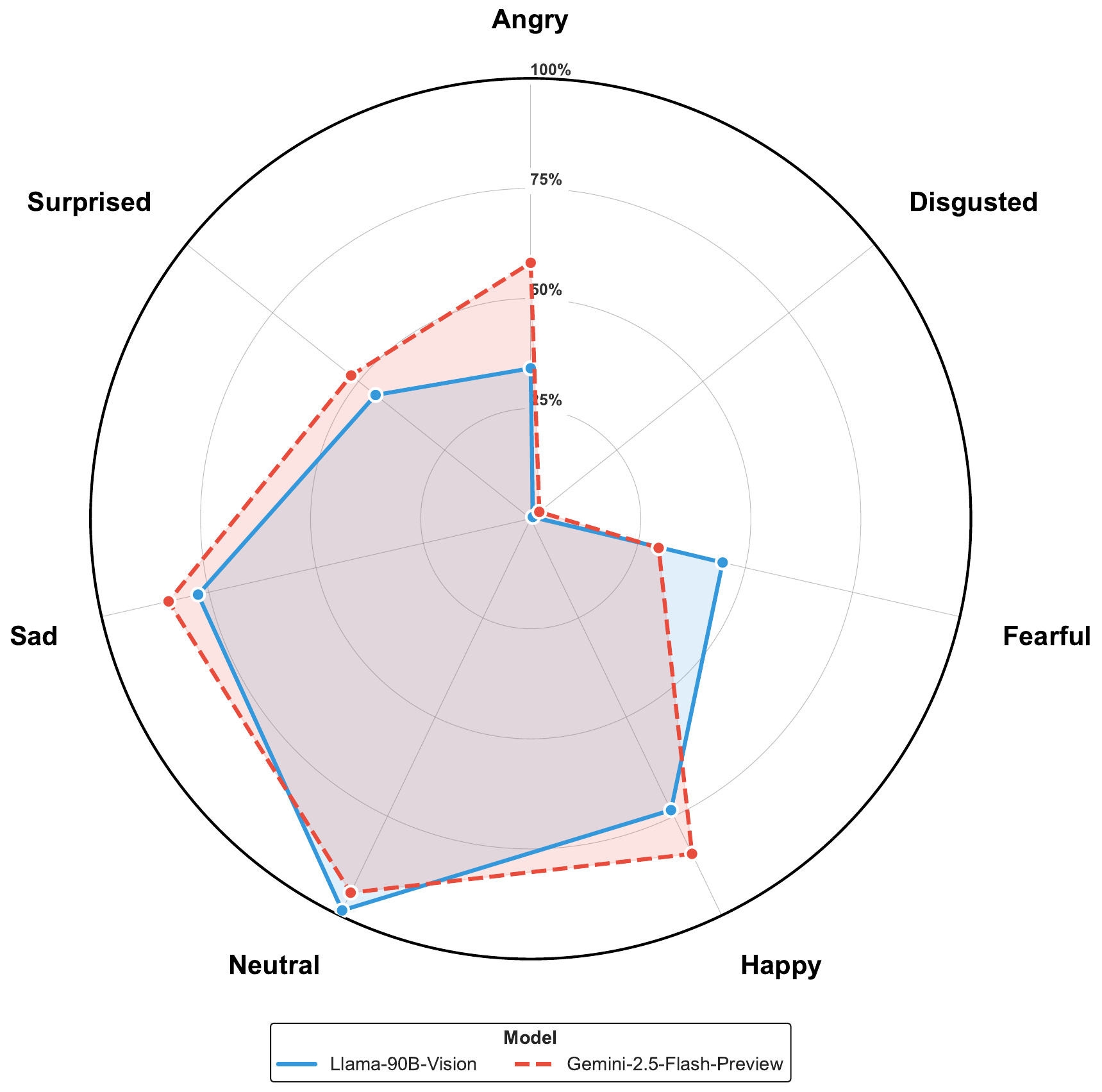}
  \caption{Radar plot of each emotion label predicted by Llama-90B and Gemini 2.5 Flash Preview model on the BESST dataset.}
  \label{fig:radar}
\end{figure}

BESST consists of staged emotional expressions in controlled visual contexts, with each emotion category enacted in both frontal and averted views (Figure~\ref{fig:besst}). To analyze how vision-language models interpret embodied emotion, we employ additional models, namely Llama-90B-Vision, Janus-7B \cite{chen2025janusprounifiedmultimodalunderstanding}, and Qwen 2.5 \cite{wang2024qwen2}, to compare their performance with the previous best results by Gemini-2.5-Flash Preview on the BESST dataset. We also tested several other LVLMs on BESST. Despite identical prompting structures, the models diverged in their calibration of embodied emotion. As shown in Table~\ref{tab:besst_performance}, Gemini 2.5 Flash outperforms all other models tested on BESST, achieving the highest F1 score (52.7), followed by Llama-90B (49.3). These differences are consistent with qualitative trends in prediction diversity, suggesting that Gemini’s more extensive pretraining and model capacity may support greater emotion generalization in embodied contexts. 
Figure~\ref{fig:radar} visualizes the distribution of predicted emotion labels across the best two models. Both models heavily favored neutral and sadness-related predictions, with notably lower frequencies for anger, disgust, and fear. This may indicate a model-level tendency to downplay high-arousal or negative valence expressions, particularly when bodily cues are ambiguous. Notably, Gemini 2.5 Flash showed a broader emotional spread, selecting anger and surprise more frequently than Llama-90B, suggesting heightened sensitivity to expressive gestures.

\section{Appendix: Emotion Mapping and Dataset Quality Analysis}
\label{sec:addres}

\begin{table}[t]
\centering
\setlength{\tabcolsep}{3.5pt}
\begin{tabular}{l|cccc}
\toprule
\textbf{Model} & \textbf{Acc} & \textbf{Prec} & \textbf{Rec} & \textbf{F1} \\
\midrule
Gemma-3-12B & 46.9 & 55.5 & 46.9 & 41.6 \\
Llama-3.2-11B & 44.8 & 48.3 & 44.9 & 37.8 \\
Llama-3.2-90B & 53.4 & 65.81 & 53.5 & 49.3 \\
\textbf{Gemini-2.5-Flash} & 57.9 & 64.8 & 58.0 & \textbf{52.7} \\
Qwen2.5-VL-7B & 42.7 & 41.2 & 42.7 & 42.4 \\
Janus-Pro-7B & 28.3 & 53.7 & 28.2 & 28.7 \\
\bottomrule
\end{tabular}
\caption{Performance comparison of large vision-language models on the BESST dataset. Scores are macro-averaged.}
\label{tab:besst_performance}
\end{table}

We also implement a two-stage pipeline, i.e., description generation followed by emotion parsing. We do so by first generating the narrative descriptions based on the exact definition, without any interpretation of the image involved. In the second step, we feed the narratives as input to perform the emotion classification based solely on the text descriptions. This approach separates the perceptual description task from the emotion understanding task. As BESST contains images of posed emotions in a controlled, lab-based environment with faces already masked, it provides an ideal setting for a clean, controlled test. This allowed us to isolate the core cognitive task of interpreting unambiguous bodily cues, testing the hypothesis of whether unified or sequential processing is more effective without the confounding variables of complex backgrounds or ``in-the-wild'' ambiguity in static images.\newline
As seen in Table~\ref{tab:besst_baseline}, ELENA achieves notably higher prediction metrics than the two-stage pipeline, although the latter suffers less from rare classes being less predicted. However, the results demonstrate that unified structured generation outperforms sequential processing. We believe this is because the single-prompt approach enables the creation of more coherent predictions, as the model jointly considers embodied emotion definitions while generating both narrative descriptions and emotion labels.

\section{Appendix: YuNet Face-Masking Experiment Setup}
\label{sec:facemasking}
For this task, we utilize the quantized version of YuNet \cite{wu2023yunet} to automate face masking in images. YuNet is a fast and accurate deep learning-based face detector available for use through OpenCV. We used the block-quantized version in \textit{int-8} precision, as it has nearly the same performance as its standard release. It outputs bounding box coordinates and confidence scores for the detected faces. We then extract the bounding boxes and subsequently replace the pixel data within the region with a uniform mask color, which conceals the detected faces. We set the parameters to detect up to 20 faces, although such images occur rarely. The confidence threshold is taken to be balanced with a value of 0.5.

In short, we duplicate the dataset that masks out every person's face. We conducted this study to investigate the following LVLM characteristic: a model that relies heavily on facial features may exhibit a significant drop in metrics and produce less confident or shorter descriptions when the face is masked. On the other hand, a model that is more \textit{body-aware} might observe a smaller drop in performance.

\begin{table}[t]
\centering
\begin{tabular}{@{}lccc@{}}
\toprule
\textbf{Method } & \textbf{Precision} & \textbf{Recall} & \textbf{F1} \\
\midrule
Two-Step Baseline & 54.2 & 55.1 & 51.0 \\ 
\textbf{ELENA} & 64.8 & 58.0 & 52.7 \\
\bottomrule
\end{tabular}
\caption{Performance comparison with the two-step (description-then-parsing) baseline on the BESST dataset. All metrics are macro-averaged (\%).}
\label{tab:besst_baseline}
\end{table}

\end{document}